\documentclass[10pt,twocolumn,letterpaper]{article}
\usepackage{iccv}
\usepackage{times}
\usepackage{epsfig}
\usepackage{epstopdf}
\usepackage{graphicx}
\usepackage{enumerate}
\usepackage{amsmath}
\usepackage{amssymb}
\usepackage[ruled,vlined]{algorithm2e}
\usepackage{bbm}
\usepackage{listings}
\usepackage{subcaption}
\usepackage{kantlipsum} 
\usepackage{mwe} 

\newtheorem{theorem}{Theorem}

\iccvfinalcopy 

\def\iccvPaperID{****} 

\ificcvfinal\fi
 
\begin{document}

\title{A Batch-Incremental Video Background Estimation Model\\using Weighted Low-Rank Approximation of Matrices}

\author{Aritra Dutta\\
    King Abdullah University of Science and Technology~(KAUST)\\
    Thuwal 23955-6900, Kingdom of Saudi Arabia\\
    {\tt\small aritra.dutta@kaust.edu.sa}
    \and
    Xin Li\\
    University of Central Florida, USA\\
    4000 Central Florida Blvd, Orlando, FL-32816\\
    {\tt\small xin.li@ucf.edu}
    \and
    Peter Richt\'arik\\
    King Abdullah University of Science and Technology~(KAUST), KSA\\
    University of Edinburgh, Scotland\\
    {\tt\small peter.richtarik@kaust.edu.sa}
}

\maketitle

\begin{abstract}
Principal component pursuit~(PCP) is a state-of-the-art approach for background estimation problems. Due to their higher computational cost,~PCP algorithms, such as robust principal component analysis~(RPCA) and its variants, are not feasible in processing high definition videos.~To avoid the curse of dimensionality in those algorithms, several methods have been proposed to solve the background estimation problem in an incremental manner.~We propose a batch-incremental background estimation model using a special weighted low-rank approximation of matrices.~Through experiments with real and synthetic video sequences, we demonstrate that our method is superior to the state-of-the-art background estimation algorithms such as GRASTA, ReProCS, incPCP, and GFL.
\end{abstract}

\vspace{-0.0in}
\section{Introduction}
\vspace{-0.0in}
Background estimation and moving object detection is an important step in many computer vision systems and video-surveillance applications.~In the past decade, one of the prevalent approaches used for background estimation is to treat it as a low-rank and sparse matrix decomposition problem~\cite{Bouwmans201431,Bouwmans2016,Sobral20144}.~Oliver \etal~\cite{oliver1999} showed that when the camera motion is small, the background is not expected to change much throughout the video frames and they assumed it to be low-rank.~The seminal work of~Lin \etal, Wright \etal, and Cand\`{e}s \etal\cite{candeslimawright,LinChenMa,APG}, which is referred as robust principal component analysis~(RPCA), solves the problem of background estimation and moving object detection in a single framework.

Given a sequence of $n$ video frames with each frame ${\mathbf a}_i\in {\mathbb R}^m$ being vectorized, let the data matrix $A=({\mathbf a}_1, {\mathbf a}_2,\cdots,{\mathbf a}_n)\in {\mathbb R}^{m\times n}$ be the concatenation of all the video frames.~The foreground is usually sparse if its size is relatively small compared to the frame size~\cite{candeslimawright,LinChenMa,APG}.~Therefore, it is natural to consider a matrix decomposition problem by writing $A$ as the sum of its background and foreground:
\vspace{-0.05in}
\begin{align*}
    A=B+F,
\end{align*}
~\\[-0.2in]
where $B,F\in {\mathbb R}^{m\times n}$ are the low-rank background and sparse foreground matrices, respectively.~RPCA solves:
\vspace{-0.05in}
\begin{equation}\label{rpca}
\min_B \|A-B\|_{\ell_1}+\lambda \|B\|_*,
\end{equation}
~\\[-0.1in]
where $\|\cdot\|_{\ell_1}$ and $\|\cdot\|_*$ denote the $\ell_1$ norm and the nuclear norm~(sum of the singular values) of matrices, respetively.

Consider a situation when a few,~say $k$, principal directions are
already specified and one wants to find a rank $r$ approximation of
the data, where $k\le r$.~In 1987, Golub \etal~\cite{golub}
formulated the following constrained low-rank approximation problem~(to be referred as GHS from now on) to address this
situation:~Given $A=(A_1\;A_2)\in\mathbb{R}^{m\times n}$ with
$A_1\in\mathbb{R}^{m\times k}$ and $A_2\in\mathbb{R}^{m\times
(n-k)}$, find $A_G=(\tilde{B}_1\;\tilde{B}_2)$ such that
~\\[-0.2in]
\begin{eqnarray}
(\tilde{B}_1\;\tilde{B}_2)=\arg\min_{\substack{B=(B_1\;B_2)\\B_1=A_1\\{\rm rank}(B)\le r}}\|A-B\|_F^2,\label{golub's problem}
\end{eqnarray}
~\\[-0.12in]
where $\|\cdot\|_F$ denotes the Frobenius norm of matrices.~That is, Golub \etal~required a few columns, $A_1,$ of $A$ be preserved when looking for a low rank approximation of $(A_1\;A_2).$
When $A_1=\emptyset$, we are back to the standard problem of low-rank approximation: find $\tilde{B}$ such that
~\\[-0.2in]
\begin{eqnarray}
\tilde{B}=\arg\min_{\substack{B\\{\rm rank}(B)\le r}}\|A-B\|_F^2.\label{low-rank problem}
\end{eqnarray}
~\\[-0.12in]
As it is well known, this problem is equivalent to principal component analysis (PCA)~\cite{pca} and has a closed form solution using the singular value decomposition (SVD) of $A$: if $A=PDQ^t$ is a SVD of $A$ with unitary matrices $P,Q$ and diagonal matrix $D$ (of non-ascending diagonal entries), then the solution to (\ref{low-rank problem}) is given by $\tilde{B}=H_r(A):=PD_rQ^t$, where $D_r$ is a diagonal matrix obtained from $D$ by only keeping the $r$ largest entries and replacing the rest by $0$. The operator $H_r$ is referred to as the hard thresholding operator.~Using the
thresholding operator, GHS problem (\ref{golub's problem}) has a
closed form solution as the following theorem explains.
~\\[-0.2in]
\begin{theorem}
\label{theorem 1}\cite{golub}
 Assume ${\rm rank}(A_1)=k$ and $r\ge k$, the solution $\tilde{B}_2$ in~(\ref{golub's problem}) is given by
 ~\\[-0.2in]
    \begin{align}\label{ghs}
    \tilde{B}_2= P_{A_1}(A_2)+H_{r-k}\left(P^{\perp}_{A_1}(A_2)\right),
    \end{align}
    ~\\[-0.2in]
    where $P_{A_1}$ and $P^\perp_{A_1}$ are the projection operators to the column space of $A_1$ and its orthogonal complement, respectively.
\end{theorem}

Assuming some pure background frames are known, GHS can be applied
by using these background frames as the first block matrix $A_1$.~Along a similar line, recently, Xin \etal\cite{xin2015} proposed a
{\it supervised} learning model called generalized fused Lasso~(GFL)
which solves:
 ~\\[-0.2in]
\begin{align}\label{gfl}
\min_{\substack{B\\B=(B_1\;\;B_2)\\B_1=A_1}}{\rm rank}(B)+\|A-B\|_{gfl},
\end{align}
~\\[-0.1in]
where $\|\cdot\|_{gfl}$ denotes a norm that is a combination of the
$\ell_1$ norm and a local spatial total variation norm (to encourage
connectivity of the foreground).~To solve GFL problem (\ref{gfl}),
Xin \etal\cite{xin2015} further specialized the above model by
requiring ${\rm rank}(B)={\rm rank}(A_1)$.~Note that, with this
specialization,~problem (\ref{gfl}) can be viewed as a constrained
low-rank approximation problem as in GHS problem (\ref{golub's
problem}) and can be formulated as follows:
 ~\\[-0.12in]
\begin{equation}\label{gfl_golub}
\min_{\substack{B = (B_1\;B_2)\\{\rm rank}(B)\le r\\B_1=A_1}} \|A-B\|_{gfl}.
\end{equation}
 ~\\[-0.2in]

\vspace{-0.0in}
\subsection{Incremental Methods} 
\vspace{-0.0in}
Conventional PCA~\cite{pca} is an essential tool in numerically solving both RPCA and GFL problems.~PCA operates at a cost of $\min\{\mathcal{O}(m^2n),\mathcal{O}(mn^2)\}$ which is due to the SVD of an $m\times n$ data matrix.~For RPCA algorithms, the space complexity of an SVD computation is approximately~$\mathcal{O}((m+n)r)$, where $r$ is the rank of the low-rank approximation matrix in each iteration, which is increasing.~For a high resolution video sequence characterized by very large $m$, this results in high computational cost and memory usage for the RPCA and GFL algorithms.~For example, the accelerated proximal gradient~(APG) algorithm runs out of memory to process 600 video frames each of size $300\times 400$ on a computer with 3.1 GHz Intel Core i7-4770S processor and 8GB memory.~In the past few decades, incremental PCA~(IPCA) was developed for machine learning applications to reduce the computational complexity of performing PCA on a huge data set.~The idea is to produce an efficient SVD calculation of an augmented matrix of the form $[A\;\tilde{A}]$ using the SVD of $A$, where $A\in\mathbb{R}^{m\times n}$ is the original matrix and $\tilde{A}$ contains $r$ newly added columns~\cite{incpca}. Similar to the IPCA, several methods have been proposed to solve the background estimation problem in an incremental manner~\cite{mmb,inrpca}.~In 2012, He \etal~\cite{grasta} proposed the Grassmannian robust adaptive subspace estimation~(GRASTA), a robust subspace tracking algorithm, and showed its application in background estimation problems.~More recently,~Guo \etal~\cite{reprocs} proposed another online algorithm for separating sparse and low dimensional subspace.~Given an initial sequence of training background video frames,~Guo \etal devised a recursive projected compressive sensing algorithm~(ReProCS) for background estimation~(see also~\cite{pracreprocs,modified_cs}).~Following a modified framework of the conventional RPCA problem, Rodriguez \etal~\cite{incpcp} formulate the incremental principal component pursuit~(incPCP) algorithm which processes one frame at a time in an incremental fashion and uses only a few frames for initialization of the prior~(see also~\cite{matlab_pcp,inpcp_jitter}).~To the best of our knowledge, these are the state-of-the-art incremental background estimation models.
\vspace{-0.0in}
\subsection{Contributions}
\vspace{-0.0in}
In this paper, we propose an adaptive batch-incremental model for background estimation.~The strength of our model lies in finding the background frame indexes in a robust and incremental manner to process the entire video sequence.~Unlike the models described previously,~we do not require any training frames.~The model we use allows us to use the background information from previous batch in a natural way.

Before describing our main contribution, let us take a pause here and revisit the idea of Golub \etal.
Inspired by (\ref{golub's problem}) and motivated by applications in which $A_1$ may contain noise, it makes more sense if we require $\|A_1-B_1\|_F$ small instead of asking for $B_1=A_1$ as in~(\ref{golub's problem}). This leads Dutta \etal~\cite{duttali_acl,duttali,duttali_bg} to consider the following more general weighted low-rank~(WLR) approximation problem:
~\\[-0.2in]
\begin{eqnarray}\label{hadamard problem}
\min_{\substack{X=(X_1\;X_2)\\{\rm rank}(X)\le r}}\|\left((A_1\;A_2)-(X_1\;X_2)\right)\odot W\|_F^2,
\end{eqnarray}
~\\[-0.1in]
where $W\in\mathbb{R}^{m\times n}$ is a matrix with non-negative entires and $\odot$ denotes the Hadamard product.~Using $W=(W_1\;\mathbbm{1})$ in~\cite{duttali_acl}, the model~(\ref{hadamard problem}) was applied to solve background estimation problems.~Here we propose a batch-incremental background estimation model using the WLR algorithm of~Dutta \etal to gain robustness. Similar to the $\ell_1$ norm used in conventional and the incremental methods, the use of a weighted Frobenius norm makes WLR robust to the outliers for background estimation problems~\cite{duttali_acl,duttali_bg}.~Our batch method is fast and can deal with high quality video sequences similar to incPCP and ReProCS.~Some conventional algorithms, for example, supervised GFL or ReProCS, require an initial training sequence which does not contain any foreground object.~Our experimental results on both synthetic and real video sequences show that unlike the supervised GFL and ReProCS, our model does not require a prior instead, it can estimate its own prior robustly from the entire data. We believe the adaptive nature of the algorithm is suitable for real time high-definition video surveillance and for panning motion of the camera where the background is slowly evolving.

\begin{algorithm}
    \SetAlgoLined
    \SetKwInOut{Input}{Input}
    \SetKwInOut{Output}{Output}
    \SetKwInOut{Init}{Initialize}
    \nl\Input{$A=(A_1\;\;A_2) \in\mathbb{R}^{m\times n}$ (the given matrix), $W= (W_1\;\;\mathbbm{1})\in\mathbb{R}^{m\times n}$ (the weight), threshold $\epsilon>0$\;}
    \nl\Init {$(X_1)_0,C_0,B_0,D_0$\;}
    \nl \While{not converged}
    {
        \nl $E_p=A_1\odot W_1\odot W_1+(A_2-B_pD_p)C_p^T$\;
        \nl \For {$i=1:m$}
        {
            \nl $(X_1(i,:))_{p+1}=(E(i,:))_p({\rm diag}(W_1^2(i,1)$ $W_1^2(i,2)\cdots W_1^2(i,k))+C_pC_p^T)^{-1}$\;
        }
        \nl $C_{p+1}=((X_1)_{p+1}^T(X_1)_{p+1})^{-1}(X_1)_{p+1}^T(A_2-B_pD_p)$\;
        \nl $B_{p+1}=(A_2-(X_1)_{p+1}C_{p+1})D_p^T(D_pD_p^T)^{-1}$\;
        \nl $D_{p+1}=(B_{p+1}^TB_{p+1})^{-1}B_{p+1}^T(A_2-(X_1)_{p+1}C_{p+1})$\;
        \nl $p=p+1$\;
    }
    \nl \Output{$(X_1)_{p+1}, (X_1)_{p+1}C_{p+1}+B_{p+1}D_{p+1}.$}
    \caption{WLR Algorithm}\label{wlr}
\end{algorithm}
~\\[-0.2in]
\begin{algorithm}
    \SetAlgoLined
    \SetKwInOut{Input}{Input}
    \SetKwInOut{Output}{Output}
    \SetKwInOut{Init}{Initialize}
    \nl\Input{$p$,~$A=(A^{(1)}\;\;A^{(2)}\ldots A^{(p)}) \in\mathbb{R}^{m\times n}$,~$\tau>0$~(for SVT), $\alpha, \beta>0$~(for weights),
    threshold $\epsilon>0,$ $k_{max},~i_r\in\mathbb{N}$\;}
    \nl Run SVT on $A^{(1)}$ with parameter $\tau$ to obtain: $A^{(1)}=B^{(1)}_{In}+F^{(1)}_{In}$\;
    \nl Initialize the background block by $B=B^{(1)}_{In}$ and $A^{(0)}=A^{(1)}$\;
    \nl \For {$j=1:p$}
    {
    \nl Identify the indices $S$ of at most $k_{max}$ columns of $A^{(j-1)}$ that are closest to background using $B$ and
    $F=A^{(j-1)}-B$\;
    \nl Set $k=\#(S), r = k+i_r$\;
    \nl Set the first block:~$\tilde{A}_1= (A^{(j-1)}(:,i))_{m\times k}$ with $i\in S$\;
    \nl Define $W=(W_1\;\mathbbm{1})$ with $W_1\in\mathbb{R}^{m\times k}$ where $(W_1)_{ij}$ are randomly chosen
    from $[\alpha, \beta]$\;
    \nl Apply {\bf Algorithm 1} on $\tilde{A}^{(j)}=(\tilde{A}_1\;A^{(j)})$ using threshold $\epsilon$ and weight $W$ to obtain its
    low rank component $\tilde{B}^{(j)}$ and define $\tilde{F}^{(j)}=\tilde{A}^{(j)}-\tilde{B}^{(j)}$\;
    \nl Take the sub-matrix of $\tilde{B}^{(j)}$ corresponding to the $A^{(j)}$ block such that $A^{(j)}=B^{(j)}+F^{(j)}$\;
    \nl Update the background block: $B=\tilde{B}^{(j)}$\;
    }
    \nl \Output{$B=(B^{(1)},B^{(2)},...,B^{(p)})$.}
    \caption{Incremental Background Estimation using WLR~(inWLR)}\label{inwlr}
\end{algorithm}
~\\[-0.2in]

\vspace{-0.0in}
\subsection{The WLR algorithm}
\vspace{-0.0in}
We now give a brief overview of the WLR algorithm proposed by Dutta \etal~\cite{duttali,duttali_bg}.~Let ${\rm rank}(X_1)=k$.~Then any $X_2$ such that ${\rm rank}(X_1\;\;X_2)\le r$ can be given in the form~\\[-0.2in]
\begin{align*}X_2=X_1C+BD,~\\[-0.3in]\end{align*}
~\\[-0.1in]
for some matrices $B\in\mathbb{R}^{m\times (r-k)},$ $D\in\mathbb{R}^{(r-k)\times (n-k)},$ and $C\in\mathbb{R}^{k\times (n-k)}.$ Therefore,~problem (\ref{hadamard problem}) with $W=(W_1\;\mathbbm{1})$ of compatible block partition is reduced to:
~\\[-0.12in]
\begin{align}\label{main problem 2}
\min_{X_1,C,B,D}\|(A_1-X_1)\odot W_1\|_F^2+\|A_2-X_1C-BD\|_F^2.
\end{align}
~\\[-0.12in]
The complexity of one iteration of Algorithm~\ref{wlr} is $\mathcal{O}(mk^3+mnr)$~\cite{duttali}.

\begin{figure*}
    \centering
    \includegraphics[width=\textwidth, height = 2.6in]{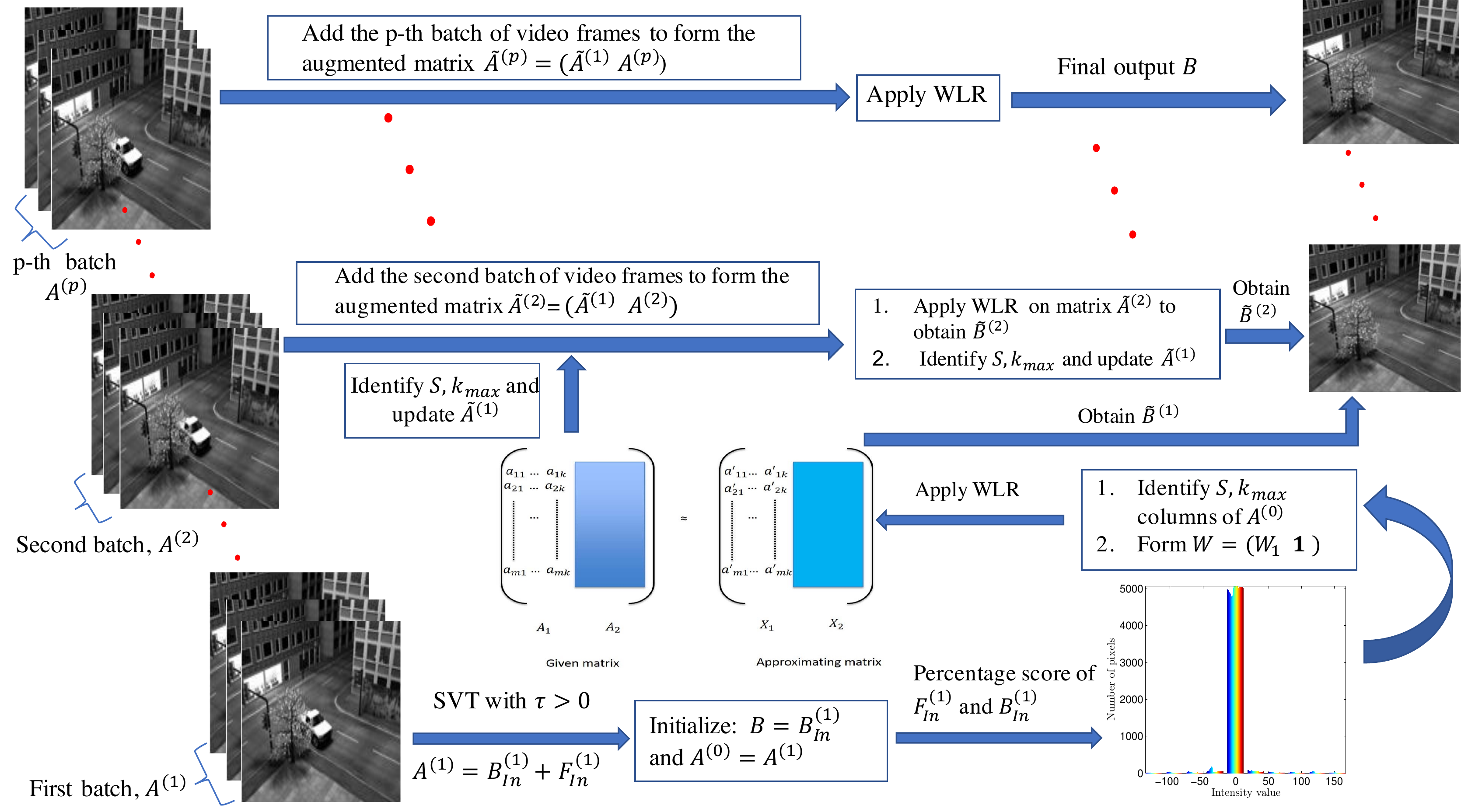}
    \caption{A flowchart for WLR inspired background estimation model proposed in Algorithm 2.}
    \label{algo_flowchart}
\end{figure*}

\vspace{-0.0in}
\section{An incremental model using WLR}
\vspace{-0.0in}
In this section, we propose an incremental weighted
low-rank approximation~(inWLR) algorithm for background estimation
based on WLR~(see Algorithm 2 and Figure~\ref{algo_flowchart}). Our
algorithm takes the full advantage of WLR in which a prior knowledge
of the background space can be used as an additional constraint to
obtain the low rank (thus the background) estimation of the data
matrix $A$. Indeed, we start by partitioning the original video
sequence into $p$ batches:~$A=(A^{(1)}\;\;A^{(2)}\ldots A^{(p)})$,
where the batch sizes do not need to be equal.~Instead of working on
the entire video sequence, the algorithm incrementally works through
each batch. To initialize, the algorithm performs a coarse
estimation of the possible background frame indices of $A^{(1)}$: we
run the classic singular value thresholding~(SVT) of Cai
\etal~\cite{caicandesshen} on $A^{(1)}$ to obtain a low rank
component (containing the estimations of background frames)
$B^{(1)}_{In}$ and let $F^{(1)}_{In}=A^{(1)}-B^{(1)}_{In}$ be the
estimation of the foreground matrix (Step 2).~From the above, we
obtain the initialization for $B$ and $A^{(0)}$ ~(Step 3). Then we
go through each batch $A^{(j)}$ using the estimates of the
background from the previous batch as prior for the WLR algorithm to
get the background $\tilde{B}^{(j)}$ (Step 9).~The identification of
the ``best background frames'' is obtained by a modified version of
the percentage score model by Dutta \etal~\cite{duttaligongshah} to
determine the indices of frames that contain the least information of
the foreground (Step 5).~This allows us to estimate $k$, $r$, and
the first block $\tilde{A}_1$ which contains the background prior
knowledge (Steps 6-7).~Weight matrix $W= (W_1\;\mathbbm{1})$ is
chosen by randomly picking entries of the first block $W_1$ from an
interval $[\alpha, \beta]$ using an uniform distribution, where
$\beta>\alpha>0$ are large (Step 8).~To understand the effect of
using a large weight in $W_1$ we refer the reader
to~\cite{duttali_acl,duttali}.~Finally, we collect background
information for next iteration (Steps 10-11).~Note that the number
of columns of the weight matrix $W_1$ is $k$ which is controlled by
bound $k_{max}$ so that the column size of $\tilde{A}^{(j)}$ is not
growing with $j$.~The output of the algorithm is the background
estimations for each batch collected in a single matrix $B$.~When
the camera motion is small, updating the first block matrix
$\tilde{A}_1$ (Step~7) has trivial impact on the model since it is
not changing much. However, when the camera is panning and the
background is continuously evolving, this could be proven very
robust as new frames are entering in the video.

\begin{figure}
    \centering
    \includegraphics[width=.5\textwidth]{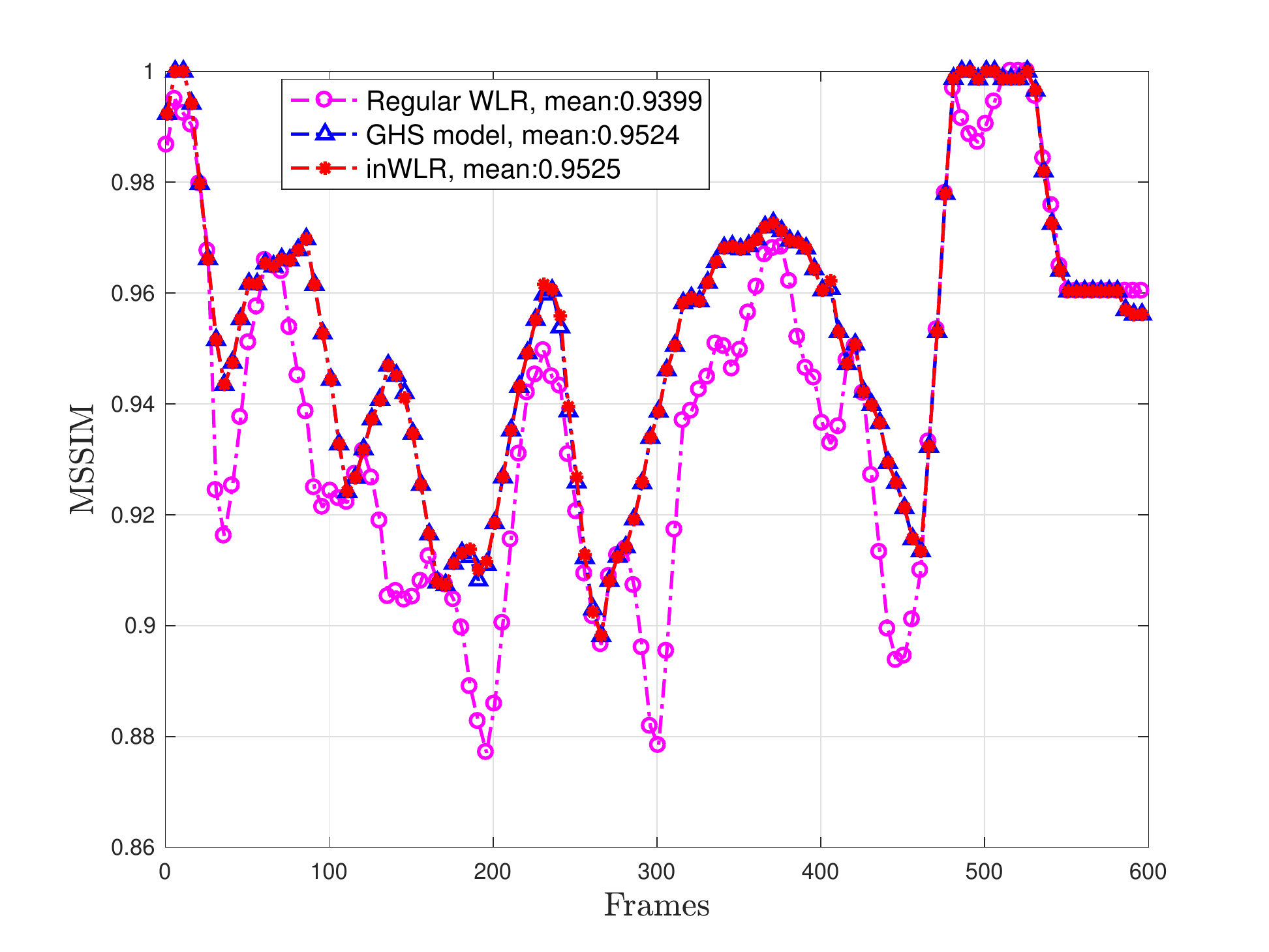}
    \caption{Comparison of MSSIM of WLR acting on all frames,~inWLR,~and GHS inspired background estimation model with frame size $[144,176]$ and $p=6$.}
    \label{ssim}
\end{figure}

\begin{figure*}
    \centering
    \includegraphics[width=\textwidth, height= 1.5in]{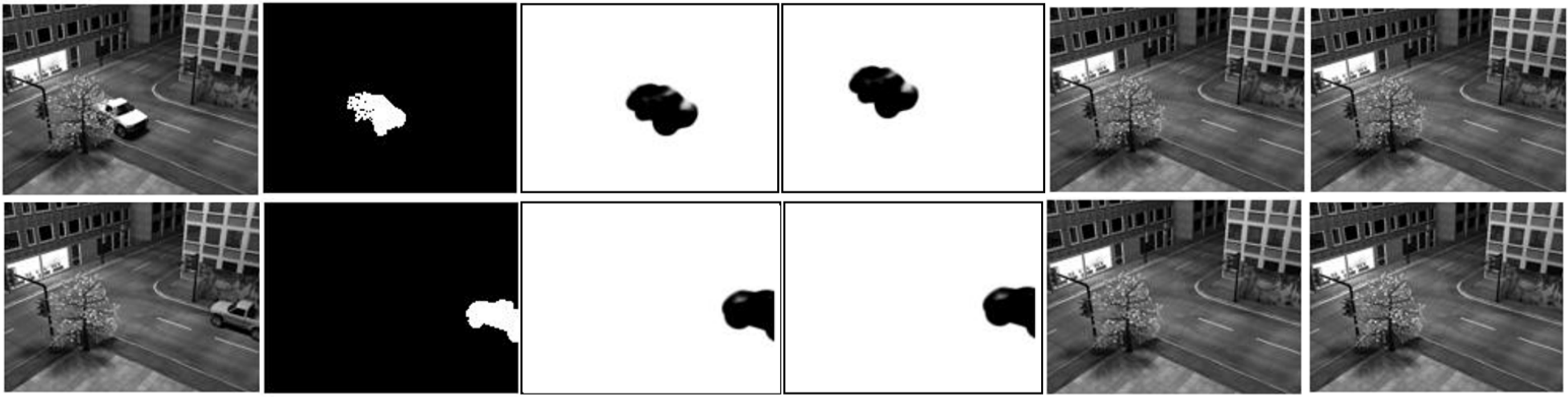}
    \caption{SSIM map of inWLR and GHS inspired background estimation model, frame size $[144,176]$, and $p=6$. Top to bottom: Frame 420 with dynamic foreground, frame 600 with static foreground.~Left to right:~Original, ground truth,~inWLR SSIM,~GHS SSIM,~inWLR background,~and~GHS background. SSIM index of the methods are 0.95027 and 0.96152, respectively.}
    \label{ssim_map}
\end{figure*}
\begin{figure*}
    \centering
    \includegraphics[width=7in,height= 1.5in]{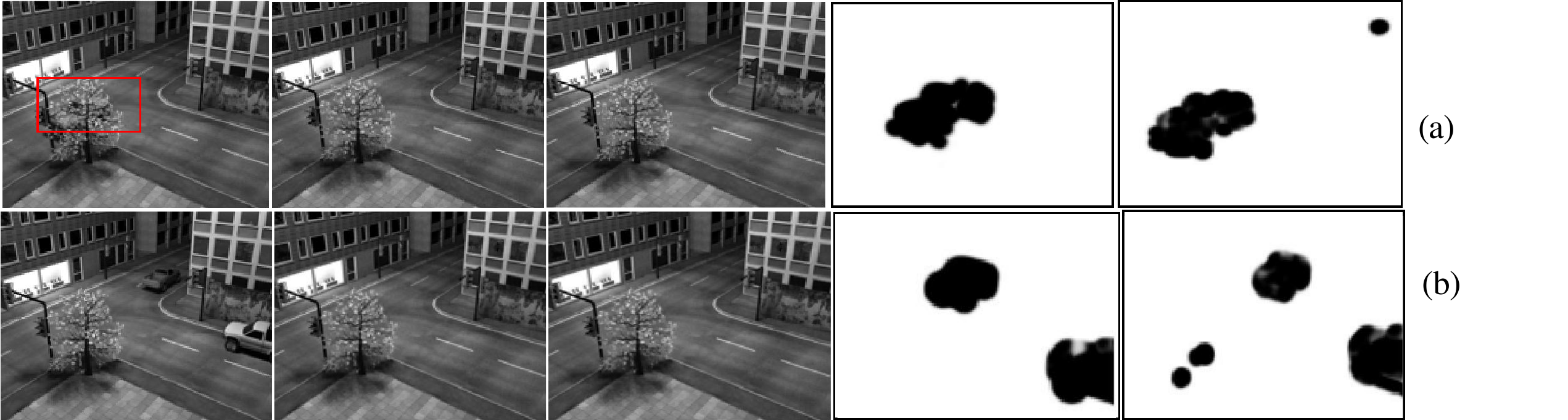}
    \caption{{\it Basic} scenario frame:~(a) 50,~(b)100.~Left to right: Original, inWLR background, GFL background,~inWLR SSIM,~and~GFL SSIM. The MSSIM of inWLR on two frames are 0.9595 and 0.9457, and that of GFL are 0.9534 and 0.9443,~respectively.}
    \label{gfl_inwlr}
\end{figure*}

\vspace{-0.0in}
\subsection{Complexity analysis}
\vspace{-0.0in}

Now,~we analyze the complexity of
Algorithm~\ref{inwlr} for equal batch size.~Primarily, the cost of
the SVT algorithm in Step~2 is only
$\mathcal{O}(\frac{mn^2}{p^2})$.~Next, in Step~9, the complexity of
implementing Algorithm~\ref{wlr}
is~$\mathcal{O}(mk^3+\frac{mnr}{p})$. Note that $r$ and $k$ are
linearly related and $k \leq k_{max}$.~Once we obtain a refined
estimate of the background frame indices $S$ as in Step~5, and, form
an augmented matrix by adding the next batch of video frames, a very
natural question in proposing our WLR inspired~Algorithm~\ref{inwlr}
is:~why do we use Algorithm~\ref{wlr} in each incremental
step~(Step~9) of Algorithm~\ref{inwlr} instead of using a closed
form solution (\ref{ghs}) of GHS? We give the following
justification:~the estimated background frames $\tilde{A}_1$ are not
necessarily exact background, only estimations of background. So,
GHS inspired model may be forced to follow the wrong data while
inWLR allows enough flexibility to find the best fit to the
background subspace. This is confirmed by our numerical experiments
(see Section~\ref{3-1} and Figure~\ref{ssim}). Thus, to analyze the
entire sequence in $p$ batches, the complexity of
Algorithm~\ref{inwlr} is approximately
$\mathcal{O}(m(k^3p+nr))$.~Note that, the complexity of
Algorithm~\ref{inwlr} is dependent on the partition $p$ of the
original data matrix. Our numerical experiments suggest for video
frames of varying sizes, the choice of $p$ plays an important role
and is empirically determined.
\begin{figure}
    \centering
    \includegraphics[width=.5\textwidth]{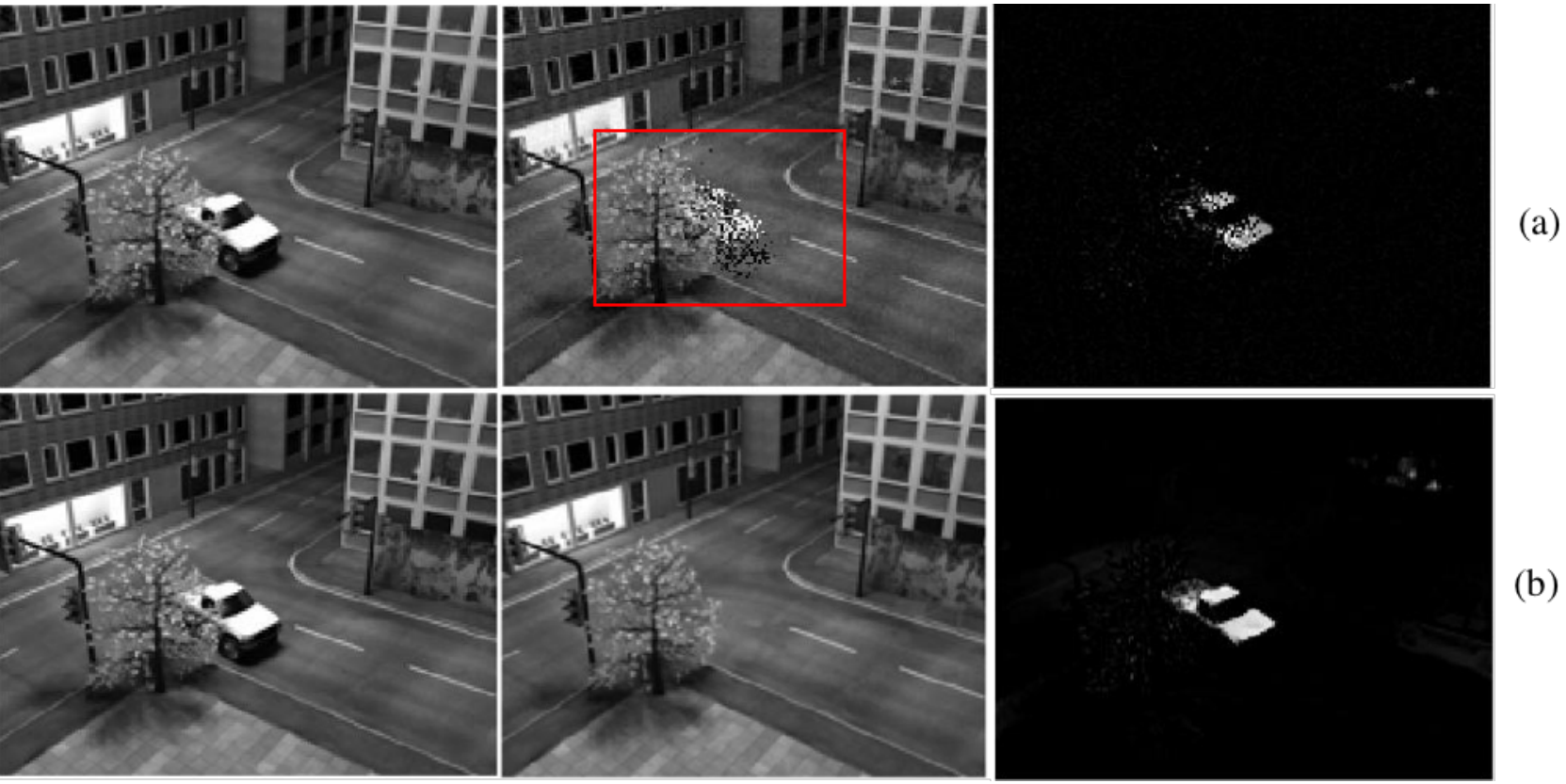}
    \caption{{\it Basic} scenario frame 420:~(a)~GRASTA,~(b)~inWLR.~Left to right:~Original, background, and foreground.~GRASTA with subsample rate 10\% recovers a fragmentary foreground and degraded background.}
    \label{grasta_qual}
\end{figure}

Unlike~Algorithm~\ref{inwlr}, if Algorithm~\ref{wlr} is used on the
entire data set, and if the number of possible background frame indices
is $k'$, then the complexity is $\mathcal{O}(m{k'}^3+mnk')$. When
$k'$ grows with $n$ and becomes much bigger than $k_{max}$ in order to achieve competitive
performance, we see that Algorithm~\ref{wlr} tends to slow down with higher
overhead than Algorithm~\ref{inwlr} does~(see Table~\ref{time}).
\vspace{-0.0in}
\section{Qualitative and quantitative analysis}
\vspace{-0.0in}
Due to the availability of ground truth frames for each foreground mask, we use 600 frames of the {\it Basic} scenario of the Stuttgart artificial video sequence~\cite{cvpr11brutzer} for quantitative and qualitative comparisons.~To capture an unified comparison against each method, we resize the video frames to [144,176] and for inWLR set $p=6$,~that is, we add a batch of 100 new video frames in every iteration until all frames are exhausted.
\begin{figure}
    \centering
    \includegraphics[width=2.7in,height = 2.1in]{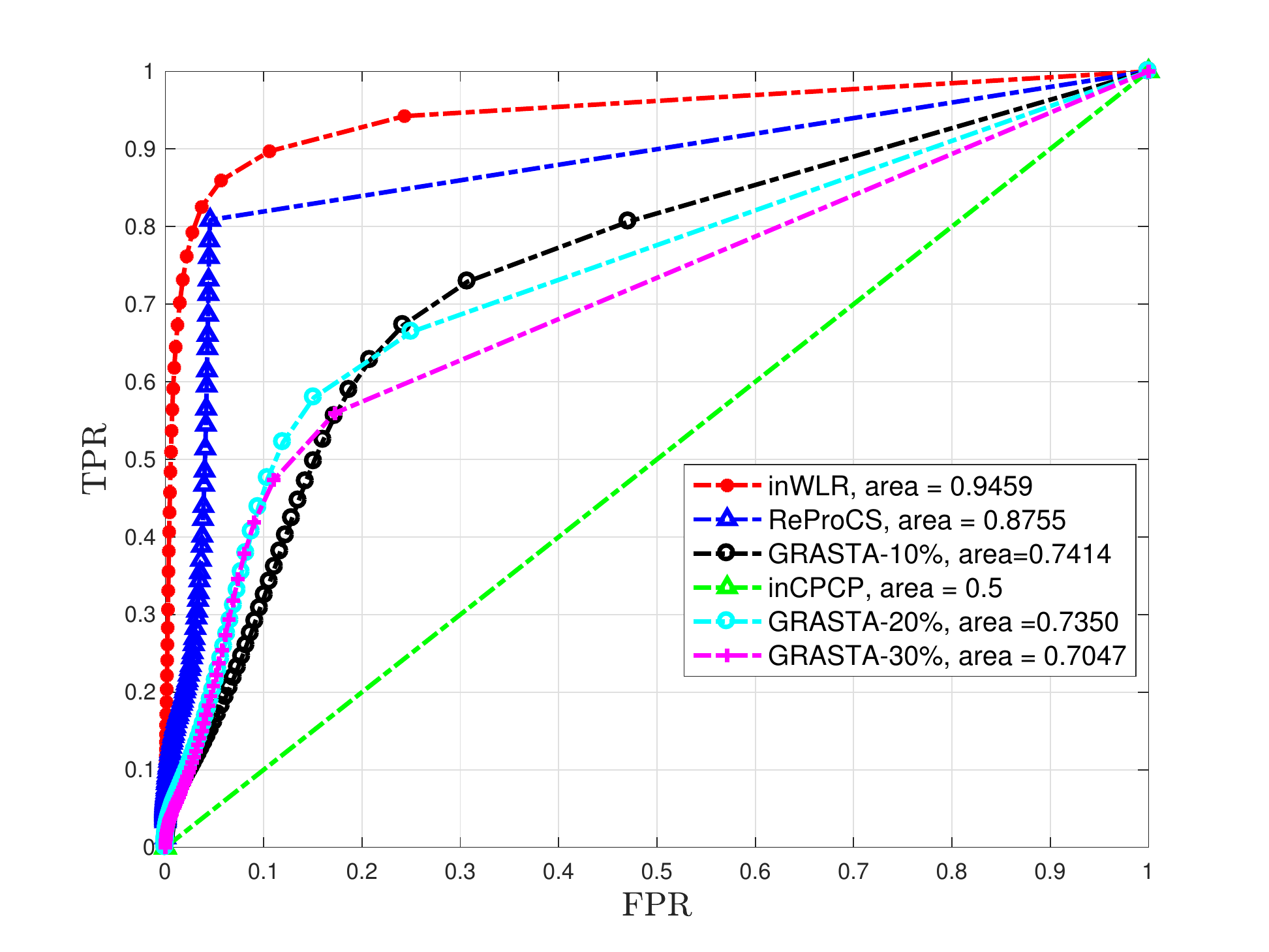}
    \caption{ROC curves on Stuttgart {\it Basic} scenario to compare between GRASTA, inWLR, incPCP, and ReProCS.}
    \label{grasta_roc}
\end{figure}
\begin{figure*}
    \centering
    \includegraphics[width=\textwidth]{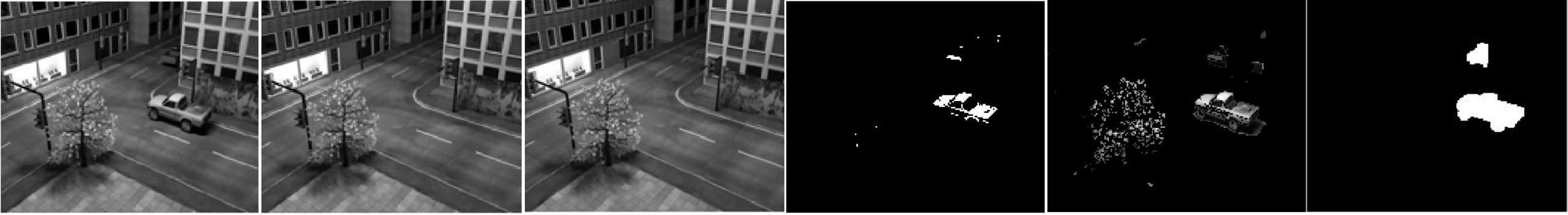}
    \caption{{\it Basic} scenario frame 123.~Left to right: Original, inWLR background, ReProCS background, inWLR foreground, ReProCS foreground,~and~ground truth.~Both methods recover similar quality background, however, ReProCS foreground has more false positives than inWLR.}
    \label{reprocs_123}
\end{figure*}
\begin{table*}
\begin{center}
\begin{tabular}{|l|c|c|c|c|c|c|c|}
\hline
Dataset & ReProCS & GRASTA & inWLR & incPCP & WLR & GHS \\
\hline
{\it Basic}& \textcolor{red}{15.8122} & 22.39 & \textcolor{blue}{17.829035} & 58.4132 & 64.0485 & 273.8382\\
\hline
{\it Fountain} &- & - & \textcolor{red}{3.709931} & - & 7.135779 &  \textcolor{blue}{4.327475}\\
\hline
{\it Waving Tree}&\textcolor{blue}{4.548824}&-&\textcolor{red}{3.3873}&-&13.751490&42.302412\\
\hline
\end{tabular}
\end{center}
\caption{Computational time comparison.~All experiments were performed on a computer with 2.7 GHz Intel Core i7 processor and 16 GB memory.~The best and the $2^{\rm nd}$ best results are colored with \textcolor{red}{red} and \textcolor{blue}{blue}, respectively. For frame numbers, frame size, and $p$ for inWLR see Section 3 and 4.}\label{time}
\end{table*}

\begin{figure*}%
\centering
\begin{subfigure}{.67\columnwidth}
\includegraphics[width=2.55in,height = 2in]{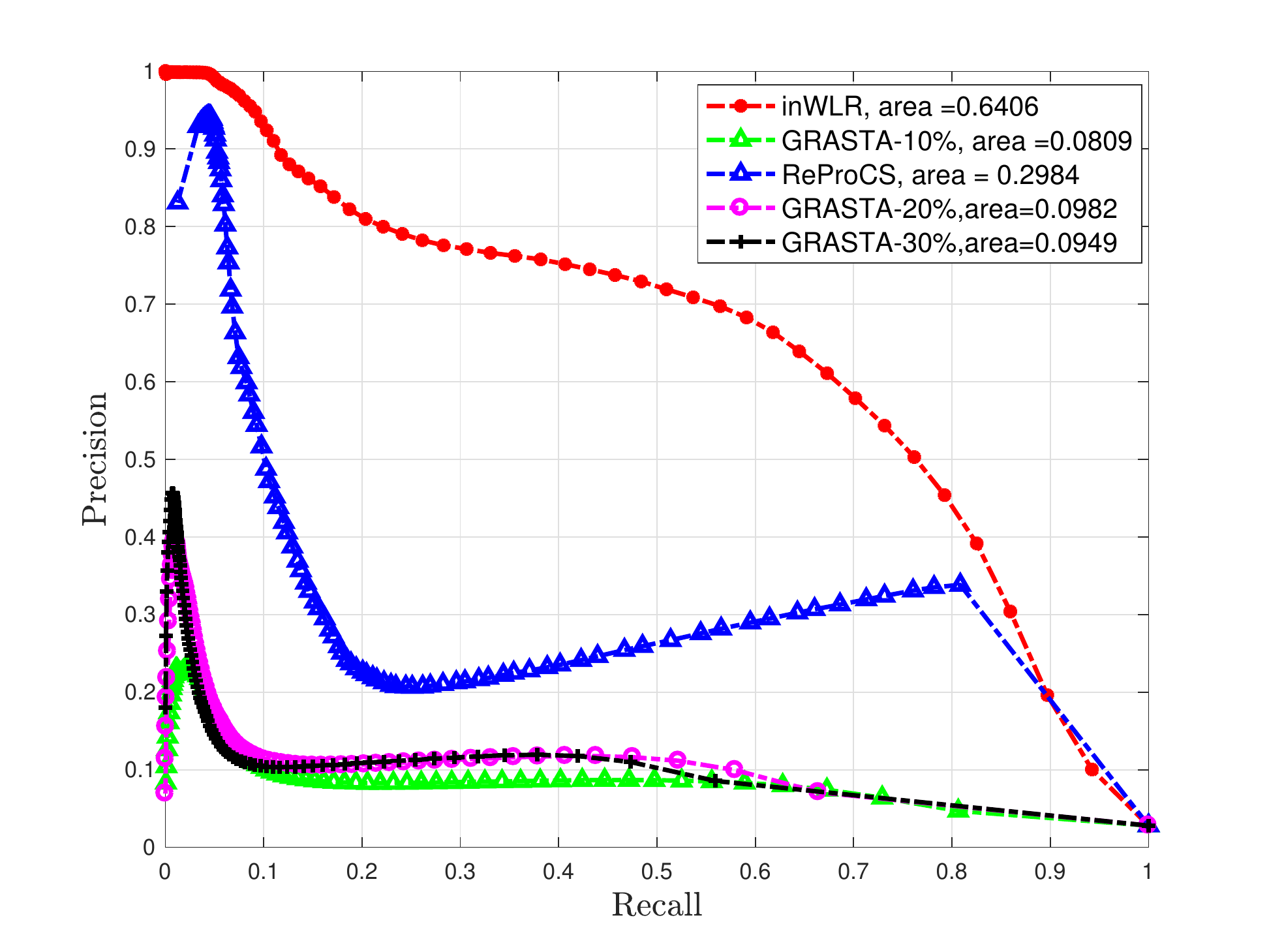}%
\caption{}%
\label{reprocs_roc}
\end{subfigure}\hfill%
\begin{subfigure}{.67\columnwidth}
\includegraphics[width=2.55in,height = 2in]{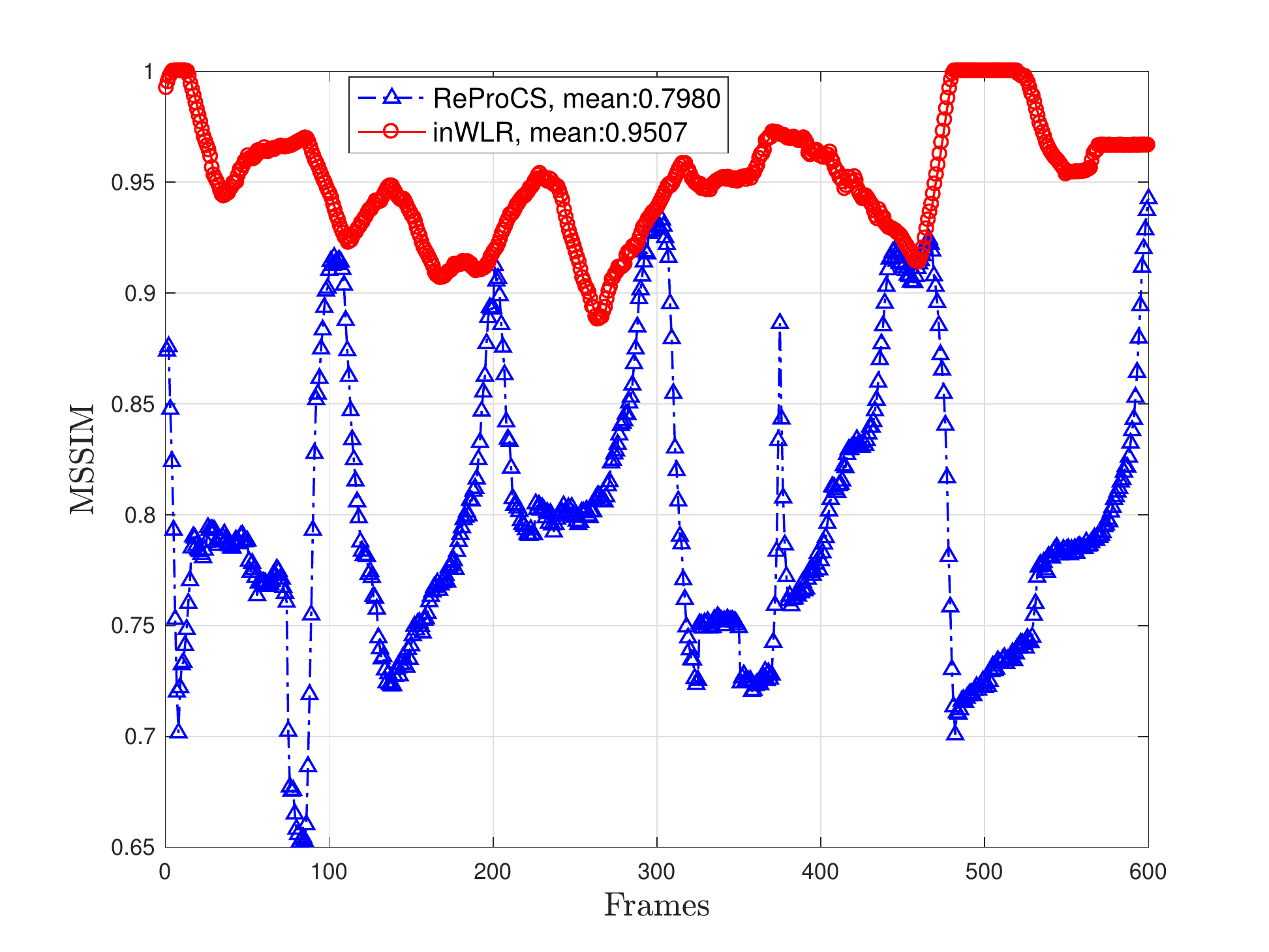}%
\caption{}%
\label{ssim_reprocs}
\end{subfigure}\hfill%
\begin{subfigure}{.67\columnwidth}
\includegraphics[width=2.55in,height = 2in]{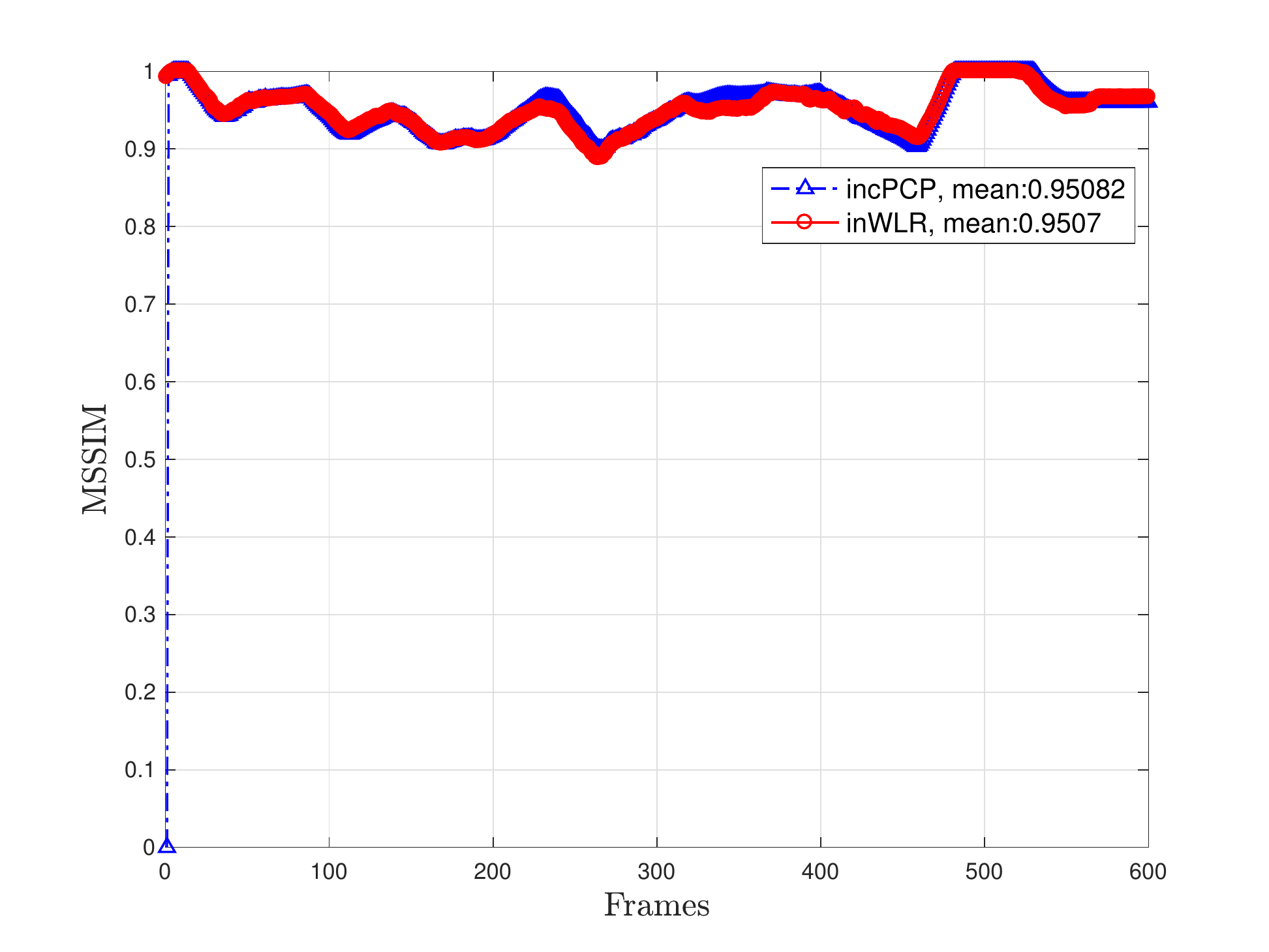}%
\caption{}%
\label{ssim_incpcp}
\end{subfigure}%
\caption{(a)~Precision-Recall curves on Stuttgart {\it Basic} scenario to compare between ReProCS, inWLR, and GRASTA.~MSSIM on Stuttgart {\it Basic} scenario to compare between:~(b)~ReProCS and inWLR,~(c)~incPCP and inWLR.}
\label{figabc}
\end{figure*}
\begin{figure}
    \centering
    \includegraphics[width=.5\textwidth]{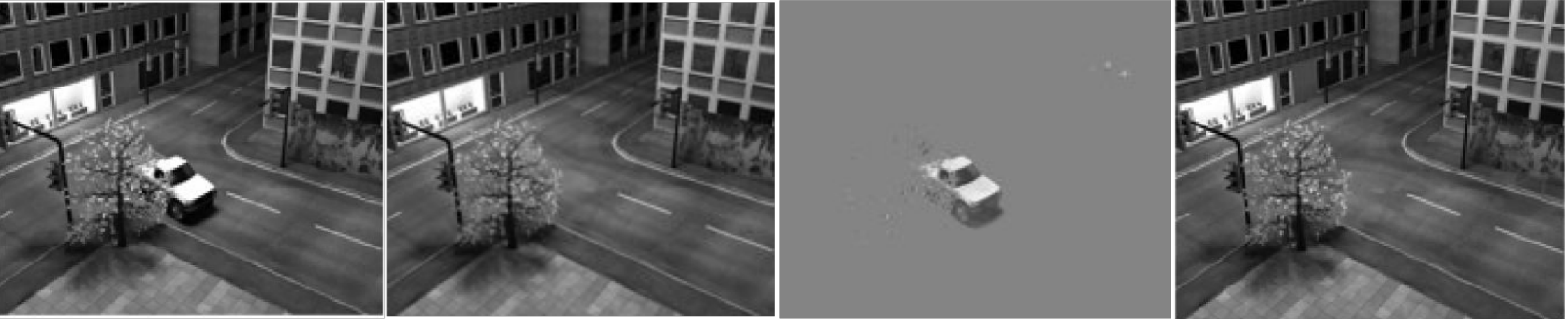}
    \caption{{\it Basic} scenario frame 420.~Left to right:~Original, incPCP background, incPCP foreground, and inWLR background. Both methods work equally well in detecting the dynamic foreground object.}
    \label{bg420_incpcp}
\end{figure}
\begin{figure*}
    \centering
    \includegraphics[width=\textwidth]{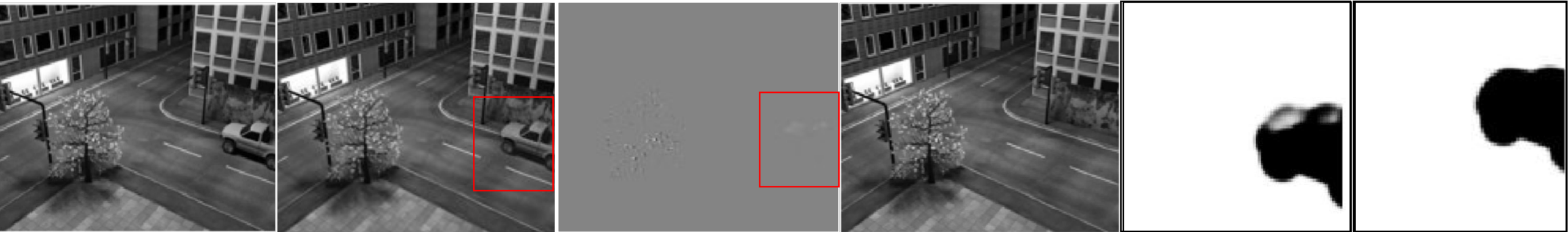}
    \caption{{\it Basic} scenario frame 600.~Left to right: Original, incPCP background, incPCP foreground, inWLR background, inWLR SSIM,~and~incPCP SSIM. incPCP fails to detect the static foreground object, though a careful reader can detect a blurry reconstruction of the car in incPCP foreground.~However, the SSIM map of both methods are equally good.}
    \label{bg600_incpcp}
\end{figure*}
\begin{figure*}
    \centering
    \includegraphics[width=\textwidth]{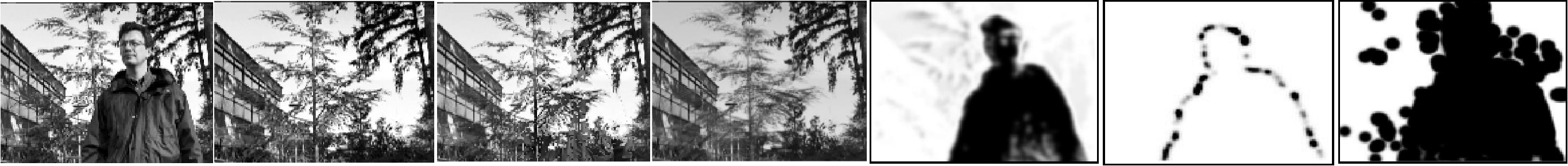}
    \caption{{\it Waving Tree} frame 247, frame size [120,160].~Left to right: Original, inWLR background~($p=6$), GFL background, ReProCS background,~inWLR SSIM~(MSSIM: 0.9592),~GFL SSIM~(MSSIM: 0.9996),~and~ReProCS SSIM~(MSSIM: 0.5221). inWLR and GFL recover superior quality background.}
    \label{waving_tree}
\end{figure*}
\vspace{-0.0in}
\subsection{Comparison with GHS}\label{3-1}
\vspace{-0.0in}
Since the~{\it Basic} scenario has no noise,~once we estimate the background frames, GHS can be used as a baseline method in comparing the effectiveness of Algorithm~\ref{inwlr}.~To demonstrate the benefit of using an iterative process as in Algorithm~\ref{wlr}, we first compare the performance of Algorithm~\ref{inwlr} against the GHS inspired models.~We also compare regular WLR acting on all 600 frames with the parameters specified in~\cite{duttali_bg}.~The structural similarity index (MSSIM) is used to quantitatively evaluate the overall image quality as it mostly agrees with the human visual perception~\cite{mssim}.~To calculate the MSSIM of each recovered foreground video frame, we consider a $11\times 11$ Gaussian window with standard deviation~($\sigma= 1.5$).~We perceive the information how the high-intensity regions of the image are coming through the noise, and consequently, we pay much less attention to the low-intensity regions.~We remove the noisy components from the foreground recovered by inWLR, $F$, by using a threshold $\epsilon_1$~(calculated implicitly in Step 5 of Algorithm 2 to choose the background frames, see~\cite{duttaligongshah}), such that we set the components below $\epsilon_1$ in $F$ to 0.~The average computation time of inWLR is approximately in the range 17.829035 seconds to 19.5755 seconds in processing 600 frames each of size $144\times 176$. On the other hand, the GHS inspired model and WLR take approximately 273.8382 and 64.5 seconds, respectively.~The MSSIM presented in Figure~\ref{ssim}, indicates that inWLR and GHS inspired model produce same result with inWLR being more time efficient than GHS.~Next in Figure~\ref{ssim_map}, the SSIM map of two sample video frames of the {\it Basic} scenario show both methods recover the similar quality background and foreground frames.~Figure~\ref{ssim} shows that to work on a high resolution video, inWLR is more accurate than GHS and WLR.
\vspace{-0.0in}
\subsection{Comparison with GFL}
\vspace{-0.0in}
We compare the performance of inWLR with the GFL
model of Xin
\etal~\cite{xin2015}\footnote{http://idm.pku.edu.cn/staff/wangyizhou/}.~For both models we use 200 frames of the {\it Basic} sequence, each
frame resized to $[144,176]$.~From the background recovered and the
SSIM map in Figures~\ref{gfl_inwlr} and \ref{waving_tree}, it is
clear that both methods are very competitive. However, it is worth
mentioning that inWLR is extraordinarily time efficient compare to
the GFL model.
\begin{figure}
    \centering
    \includegraphics[width=0.5\textwidth, height = 1.25in]{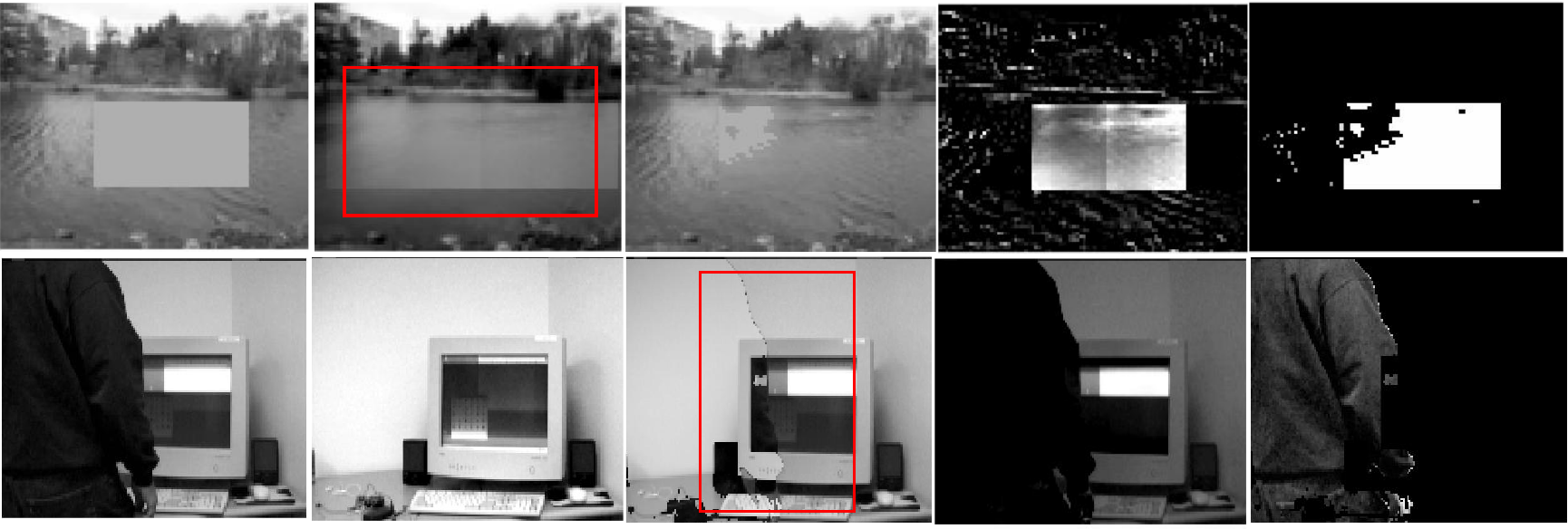}
    \caption{Left to right:~Original, inWLR background,~ReProCS background,~inWLR foreground,~and~ReProCS foreground. In {\it Lake} sequence ReProCS performs better, and in {\it Person} sequence inWLR has better performance.}
    \label{lake_person}
\end{figure}
\begin{figure}
    \centering
    \includegraphics[width=3.1in, height= 1.3in]{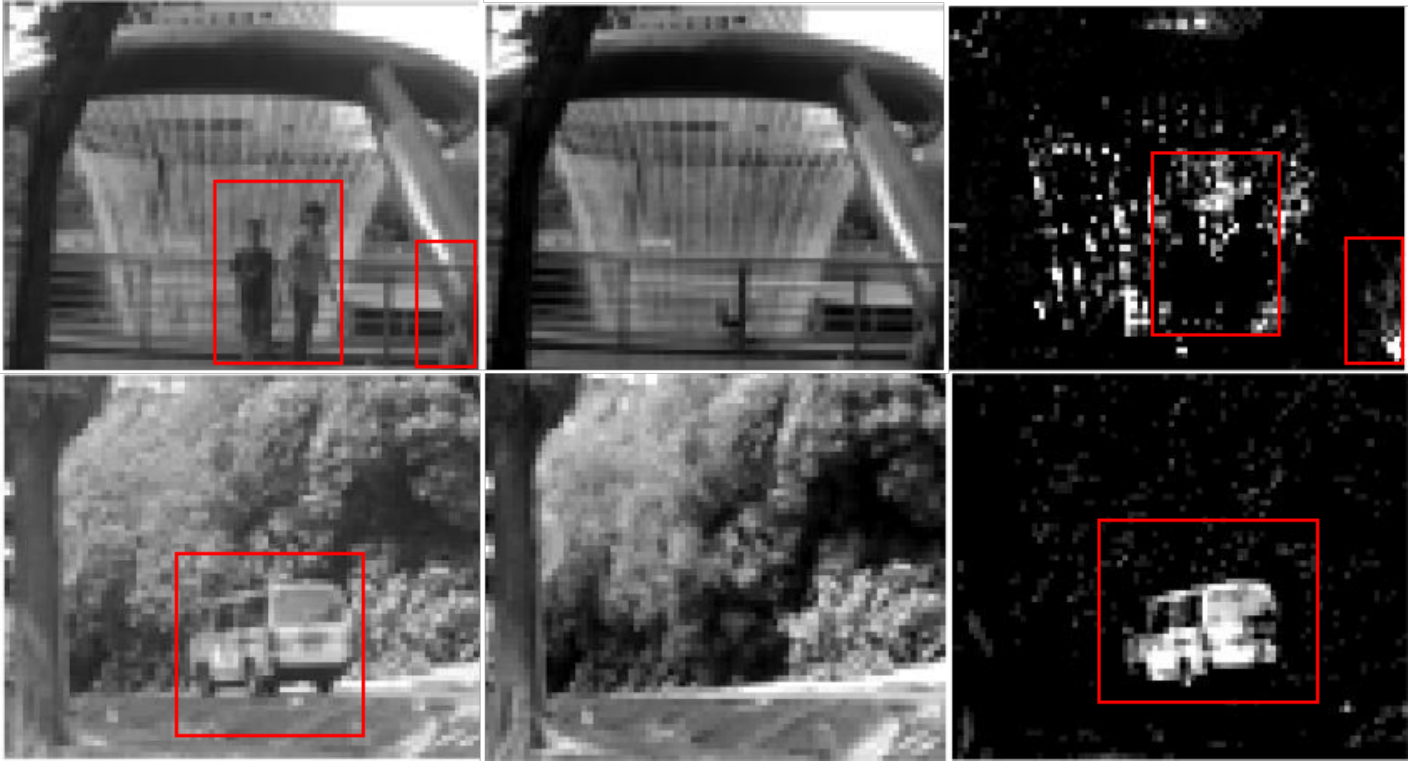}
    \caption{Top to bottom:~{\it Fountain} 500 frames with $p=5$, {\it Campus} 600 frames with $p=6$, frame size [64,80].~Left to right:~Original, inWLR background,~and~inWLR foreground.}
    \label{inwlr_bg}
\end{figure}
\vspace{-0.0in}

\vspace{-0.0in}
\subsection{Comparison with other state-of-the-art models}
\vspace{-0.00in}
In this section, we compare the performance of inWLR against other incremental background estimation models such as, GRASTA, incPCP, and ReProCS on 600 frames of the {\it Basic} scenario of the Stuttgart sequence.~For quantitative measure we use the receiver operating characteristic~(ROC) curve, the recall and precision~(RP) curve, and the MSSIM.~For ROC curve and RP curve, we use a uniform threshold vector ${\rm linspace}(0, 255,100)$ to compare pixel-wise predictive analysis between each recovered foreground frame and the corresponding ground truth frame.

\subsubsection{Comparison with GRASTA~\cite{grasta}}
\vspace{-0.01in}
At each time step $i$, GRASTA solves the following optimization problem:~For a given orthonormal basis $U_{\Omega_s}$ solve
~\\[-0.27in]
\begin{align}\label{Grasta}
\min_x\|U_{\Omega_s}x-A_{\Omega_s}(:,i)\|_{\ell_1},
\end{align}
~\\[-0.23in]
where each video frame $A(:,i)\in\mathbb{R}^n$ is subsampled over the index set ${\Omega_s}\subset\{1,2,\cdots,n\}$ following the model: $A_{\Omega_s}(:,i)=U_{\Omega_s}x+F_{\Omega_s}(:,i)+\epsilon_{\Omega_s}$,~such that, $x\in\mathbb{R}^{|{\Omega_s}|}$ is a weight vector and $\epsilon_{\Omega_s}$ is a Gaussian noise vector of same size.~After updating $x$, one has to update $U_{\Omega_s}$.~We set the subsample percentage $s$ to 0\%,~10\%,~20\%, and 30\% respectively, estimated rank 60, and keep the other parameters same as in~\cite{grasta}.~The GRSTA code is obtained from author's website.\footnote{ https://sites.google.com/site/hejunzz/grasta}~Note that, for a lower estimated rank GRASTA does not perform well.~Referring the qualitative result in Figure~\ref{grasta_qual}, we only provide the ROC curve and RP curve to compare GRASTA with different subsamples $s$ and inWLR~(see Figure~\ref{grasta_roc} and~\ref{reprocs_roc}).~The ROC curves and RP curves clearly show the superior performance of inWLR on the Stuttgart {\it Basic} scenario.

\vspace{-0.in}
\subsubsection{Comparison with ReProCS~\cite{reprocs}}
\vspace{-0.0in}
ReProCS is a two stage algorithm.~In the first stage,~given a sequence of training background frames, say $t$, the algorithm finds an approximate basis which is ideally of low-rank.~After estimating the initial low-rank subspace, in the second stage, the algorithm recursively estimates $F_{t+1}, B_{t+1}$, and the subspace in which $B_{t+1}$ lies.~We use 200 background frames of the {\it Basic} sequence for initialization of ReProCS. Figure~\ref{reprocs_123} shows both methods recover similar quality background.~However, ReProCS foreground contains more false positives than inWLR foreground.~The ROC curve, RP curve, and MSSIM in Figure~\ref{grasta_roc},~\ref{reprocs_roc},~and~\ref{ssim_reprocs} comply with our claim quantitatively for the {\it Basic} sequence. Though the average computation time for ReProCS is 15.644460 seconds which is better than inWLR.
\vspace{-0.0in}
\subsubsection{Comparison with incPCP~\cite{incpcp}}
\vspace{-0.0in}
incPCP follows a modified framework of PCP but with the assumption that the partial rank $r$ SVD of first $k-1$ background frames $B_{k-1}$ is known and using them $A_{k-1}$ can be written as $A_{k-1}=B_{k-1}+F_{k-1}$.~Therefore for a new video frame $A(:,k)$ one can solve the optimization problem as follows:
~\\[-0.1in]
\begin{align*}\label{reprocs_problem}
&\min_{\substack{B_k,F_k\\{\rm rank}(B_k)\le r}}\|B_k+F_k-A_k\|_F^2+\lambda\|F_k\|_{\ell_1},
\end{align*}
~\\[-0.15in]
where $A_k = [A_{k-1}~~A(:,k)]$ and $B_k = [U_r\Sigma_rV_r^T~~B(:,k)]$ such that $U_r\Sigma_rV_r^T$ is a partial SVD of $B_{k-1}.$ According to~\cite{incpcp}, the initialization step can be performed incrementally. For the Stuttgart sequence, the algorithm uses the first video frame for initialization.~The incPCP code is downloaded from author's website\footnote{https://sites.google.com/a/istec.net/prodrig/Home/en/pubs/incpcp}.~From the MSSIM presented in Figure~\ref{ssim_incpcp} and the background recovered by both methods in Figure~\ref{bg420_incpcp}, it seems that both methods perform equally well on the {\it Basic} scenario.~However, when the foreground is static (as in frames 551-600 of the Stuttgart sequence), the~$\ell_1$ norm in incPCP is unable to capture the foreground object, resulting the presence of the static car as a part of the background~(see Figure~\ref{bg600_incpcp}).~On the other hand, our inWLR successfully detects the static foreground.

\vspace{-0.1in}
\section{Results on real world sequences}
\vspace{-0.07in}
 In this section, we demonstrate the performance of
inWLR on five challenging real world video
sequences~\cite{lidata,wallflower}, containing occlusion, dynamic
background, and static foreground.~In Figure~\ref{waving_tree} we
compare inWLR against GFL and ReProCS on 60 frames of {\it Waving Tree} sequence.~ReProCS and GFL use 220 and
200 pure background frames respectively as training data.~In
Figure~\ref{lake_person} we compare inWLR only against ReProCS on
two complex sequences:~80 frames of {\it Lake}, frame size $[72,90]$,~and,~50 frames of {\it Person}, frame size $[120,160]$. In those sequences, for inWLR, we set $p=8$ and $5$, respectively.~Due to the
absence of ground truth we only provide qualitative comparison.~In
Figure~\ref{inwlr_bg} we demonstrate the performance of inWLR on two
data sets with dynamic background and semi-static foreground.~In
almost every video sequence, inWLR performs reasonably well.~See
Table~\ref{time} for the comparisons between computational time.

\vspace{-0.11in}
\section{Conclusion}
\vspace{-0.04in}
In this paper we propose a novel background
estimation model which operates on the entire data set in a
batch-incremental way and adaptively determines the background
frames without requiring any prior estimate.~The proposed model
demands less on storage and allows slow change in
background.~Through extensive qualitative and quantitative
comparison on real and synthetic video sequences, we establish our
claim and demonstrate the robustness of our model.~The batch sizes
and the parameters in our model are still empirically selected.~In
future, we plan to propose a more robust estimate of the parameters
and explore the possibilities in dealing with videos of more dynamic
background using our algorithm.


{\small
\bibliographystyle{ieee}
\bibliography{egbib}
}

\end{document}